\documentclass[conference]{IEEEtran}
\IEEEoverridecommandlockouts
\usepackage{cite}
\usepackage{amsmath,amssymb,amsfonts}
\usepackage{hyperref}
\usepackage{algorithmic}
\usepackage{graphicx}
\usepackage{textcomp}
\usepackage{makecell}
\usepackage{xcolor}
\usepackage[shortlabels]{enumitem}
\def\BibTeX{{\rm B\kern-.05em{\sc i\kern-.025em b}\kern-.08em
    T\kern-.1667em\lower.7ex\hbox{E}\kern-.125emX}}
\begin{document}

\bstctlcite{IEEEexample:BSTcontrol}

\title{PowerGridworld: A Framework for Multi-Agent Reinforcement Learning in Power Systems\\
\thanks{All authors are with the National Renewable Energy Laboratory, Golden, CO 80401, USA.  Corresponding author:  D. Biagioni (dave.biagioni@nrel.gov). This work was authored by the National Renewable Energy Laboratory (NREL), operated by Alliance for Sustainable Energy, LLC, for the U.S. Department of Energy (DOE) under Contract No. DE-AC36-08GO28308. This work was supported by the Laboratory Directed Research and Development (LDRD) Program at NREL. The views expressed in the article do not necessarily represent the views of the DOE or the U.S. Government. The U.S. Government retains and the publisher, by accepting the article for publication, acknowledges that the U.S. Government retains a nonexclusive, paid-up, irrevocable, worldwide license to publish or reproduce the published form of this work, or allow others to do so, for U.S. Government purposes.}
}

\author{David Biagioni, Xiangyu Zhang, Dylan Wald, Deepthi Vaidhynathan,\\Rohit Chintala, Jennifer King, Ahmed S. Zamzam}

\maketitle

\begin{abstract}
We present the PowerGridworld software package to provide users with a lightweight, modular, and customizable framework for creating power-systems-focused, multi-agent Gym environments that readily integrate with existing training frameworks for reinforcement learning (RL).  Although many frameworks exist for training multi-agent RL (MARL) policies, none can rapidly prototype and develop the environments themselves, especially in the context of heterogeneous (composite, multi-device) power systems where power flow solutions are required to define grid-level variables and costs.  PowerGridworld is an open-source software package that helps to fill this gap.  To highlight PowerGridworld's key features, we present two case studies and demonstrate learning MARL policies using both OpenAI's multi-agent deep deterministic policy gradient (MADDPG) and RLLib's proximal policy optimization (PPO) algorithms. In both cases, at least some subset of agents incorporates elements of the power flow solution at each time step as part of their reward (negative cost) structures.
\end{abstract}


\section{Introduction}

\subsection{Multi-Agent Reinforcement Learning in Energy Systems}
With the increased controllability of power consumption and generation at the edge of modern power systems, devising centralized control approaches to manage flexible devices within these systems is becoming nearly impossible. In particular, the dynamical models of these interconnected systems present considerable nonlinearities, which challenges the applicability of classical control methods. In addition, the conflicting operational costs/objectives of heterogeneous devices often obstruct the formulation of a system-wide operational objective. Thus, decentralized, data-driven control approaches, where edge controllers utilize data to derive effective local control policies, provide a viable pathway to realizing the resilient and efficient operation of future energy systems.

\begin{figure}[t]
\centering
\includegraphics[width=1.0\linewidth]{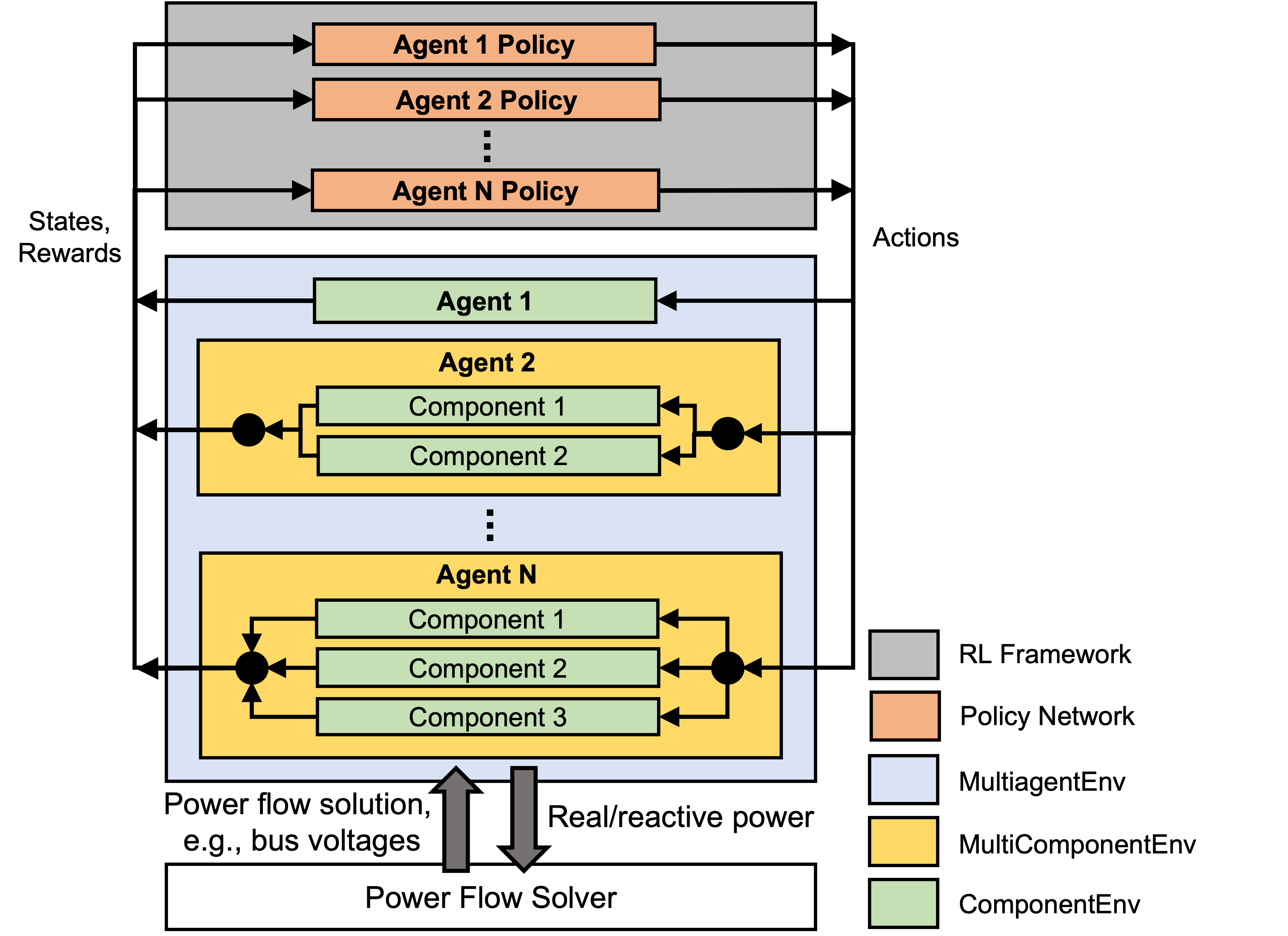}
\caption{PowerGridworld architecture for an $N$-agent environment comprised of both single-component agents (Agent 1) and multi-component agents, which can be added in any number and combination.  Given a base system load and the agents' individual power profiles, a power flow solution is computed at each control step and may be used to update the agents' states and rewards.}
\label{fig-architecture}
\end{figure}

Reinforcement learning (RL) approaches have shown great potential in several power systems control and load management tasks \cite{claessens2016convolutional, xu2019optimal, yang2019two, duan2019deep}. In addition, multi-agent RL (MARL) approaches have advanced and have been applied in many complex systems, including games \cite{vinyals2019grandmaster} and autonomous driving \cite{shalev2016safe}. Recently, MARL approaches have also found applications in the power systems domain, with an emphasis on voltage regulation problems \cite{gao2021consensus, chen2021powernet, pigott2021gridlearn}. These applications utilize the capabilities of MARL to devise local control policies without any knowledge of the models of the underlying complex systems. However, due to the significant focus on applying MARL in power system decentralized control tasks, there has been no standardized RL test environment that supports the development of heterogeneous power system components and the deployment of off-the-shelf MARL approaches. 

\subsection{Related Software}

Table \ref{tab-software} summarizes the features of MARL frameworks for power systems.

\begin{table*}[t]
  \centering
  \caption{Comparison of software features for MARL environments for power systems.}
  \label{tab-software}
  \begin{tabular}{l|p{2.4cm}|p{1.9cm}|p{1.5cm}|p{2cm}|p{3.5cm}}
    \Xhline{4\arrayrulewidth}
    \textbf{Package} & \textbf{Power Systems \qquad Application} & \textbf{Agent Customization} & \textbf{Composable Agents} & \textbf{Control Step} & \textbf{MARL Training Interfaces}\\
    \Xhline{4\arrayrulewidth}
    PettingZoo & None & Unlimited  & No & User-defined & RLLib, OpenAI \\
    \hline
    CityLearn & Demand response & Buildings and subsystems  & No & 1-hour & Package-specific \\
    \hline
    GridLearn & Demand response, voltage regulation & Buildings and subsystems  & No & Sub-hourly & Package-specific \\
    \hline
    PowerGridworld & Any energy management/optimization & Unlimited & Yes & User-defined & RLLib, OpenAI\\
    \hline
  \end{tabular}
  \vspace{-5pt}
\end{table*}

\subsubsection{General MARL Framework}

The development of MARL is relatively new compared to its single-agent counterpart, and there is currently no commonly and widely used MARL framework. PettingZoo \cite{terry2020pettingzoo}, a Python library, has a goal of developing a universal application programming interface (API) for formulating MARL problems, as OpenAI Gym \cite{brockman2016openai} did for single-agent RL problems. However, because the key advantages of PettingZoo---namely, an efficient formulation suitable for the turn-based games and an ability to handle agent creation and death within episodes---are less relevant to power system control problems, PowerGridworld does not adopt the PettingZoo APIs at this stage for simplicity. 

\subsubsection{Multi-Agent Energy Systems}
CityLearn \cite{vazquez2019citylearn,vazquez2020citylearn} is an open-source library aimed at implementing and evaluating MARL for building demand response and energy coordination.  By design, the heating and cooling demands of buildings in CityLearn are guaranteed to be satisfied, allowing researchers to focus on energy balance and load shifting for the control problem.  To achieve this, building thermal models and associated energy demands are precomputed using EnergyPlus \cite{crawley2001energyplus}, and control actions are limited to active energy storage decisions rather than those affecting passive (thermal) mass.  CityLearn is intended to provide a benchmark MARL environment from the standpoint of building demand response and, as such, it is highly constrained in terms of the types of agents and models that are available.  CityLearn energy models include buildings, heat pumps, energy storage, and batteries, while the state and action spaces of the agents themselves must be constructed from a predefined variable list. Control steps in CityLearn are restricted to 1-hour resolution, and grid physics is not modeled.
To address this, GridLearn \cite{pigott2021gridlearn} utilizes the building models provided in CityLearn and extends its functionality to include power system simulation. The added power flow model, implemented using pandapower \cite{thurner2018pandapower}, allows researchers studying decentralized control to consider both building-side and grid-level objectives. The GridLearn case study presented in \cite{pigott2021gridlearn} demonstrates that this platform can be used to train MARL controllers to achieve voltage regulation objectives in a distribution system by controlling behind-the-meter resources.


\subsubsection{MARL Training}
While many open-source choices exist for MARL training, we highlight two of the most popular:  RLLib (multiple algorithms available) and OpenAI's multi-agent deep deterministic policy gradient (MADDPG). RLLib \cite{liang2018rllib,rllib-webpage} is a framework for scalable RL training built on the Ray Python library \cite{moritz2018ray}, and it supports a variety of training paradigms for single-agent, multi-agent, hierarchical, and offline learning. RLLib can be deployed on both cloud and high performance computing (HPC) systems, and it provides a number of training ``abstractions," enabling users to develop custom, distributed RL algorithms.  The multi-agent API in PowerGridworld is derived from RLLib's own \texttt{MultiAgentEnv} API and thus is readily integrated into this framework. OpenAI\cite{openai} has played a central role in the evolution of both theory and software for RL and MARL.  In addition to creating the Gym API, OpenAI released a series of tutorials and implementations in the mid-2010s that have continued to hold traction in the RL community, including the SpinningUp blog\footnote{\url{https://spinningup.openai.com/en/latest/}} and the baselines GitHub repository.\footnote{https://github.com/openai/baselines}  The OpenAI implementation \cite{maddpg-webpage} of the MADDPG \cite{lowe2017multi} is a popular choice for MARL with continuous control.

As described in greater detail in Section \ref{sec-software-description}, PowerGridworld makes it easy for users to leverage the implementations of both RLLib's and OpenAI's RL algorithms.

To the best of our knowledge, no previous software packages exist that enable users to implement arbitrary multi-agent scenarios with with a power systems focus---in particular, with the ability to incorporate power flow solutions into the agents' observation spaces and rewards.  We believe that PowerGridworld begins to bridge this gap by enabling highly modular, customizable environments that readily integrate with open-source, scalable MARL training frameworks such as RLLib.

\section{PowerGridworld}

\subsection{Description of Software}\label{sec-software-description}
PowerGridworld is designed to provide users with a lightweight, modular, and  customizable framework for creating power-systems-focused, multi-agent Gym environments that readily integrate with existing RL training frameworks.  The purpose of this software, which is available as an open-source Python package\footnote{\href{https://github.com/NREL/PowerGridworld}{\url{https://github.com/NREL/PowerGridworld}}}, is to enable researchers to rapidly prototype simulators and RL algorithms for power systems applications at the level of detail of their choice, while also enabling the use of cloud and HPC via integration with scalable training libraries such as RLLib \cite{liang2018rllib} and Stable Baselines \cite{raffin2019stable}.  

\subsubsection{Architecture} The PowerGridworld design pattern is based on the OpenAI Gym API, which has become the \emph{de facto} standard interface for training RL algorithms.  The Gym API essentially consists of the following two methods:
\begin{itemize}
    \item \texttt{reset}:  Initialize the simulation instance and return an observation of the initial state space, $s_0$.
    \item \texttt{step}:  For each control step, apply an input control action, $a_t$, and return a a new state space observation, $s_t$; a step reward, $r_t$; a termination flag; and any desired metadata.
\end{itemize}
A simulator that is wrapped in the Gym API is often referred to as an \emph{environment}, and one instance of the simulation is often called an \emph{episode}.

The core functionality of the PowerGridworld package is to extend the Gym API to include multi-agent simulations and to allow a user to combine environments that simulate individual devices or subsystems into a single, multi-component agent.  This ``plug-and-play" functionality is highly useful in power systems applications because it enables the user to rapidly create heterogeneous agents using basic building blocks of distributed energy resources (DERs).  

We illustrate the PowerGridworld architecture in Fig. \ref{fig-architecture}.  Here, the \texttt{MultiAgentEnv} environment (blue) encapsulates $N$ agents that subclass one of two types:
\begin{enumerate}[a)]
    \item \texttt{ComponentEnv} environments (green), which implement a single, independent agent.  This class is a slight extension of the OpenAI Gym API.
    \item \texttt{MultiComponentEnv} environments (yellow), which are a composition of component environments.  For example, Agent $N$ could represent a smart building agent composed of building thermodynamics, photovoltaics (PV), and battery physics, each implemented as a separate \texttt{ComponentEnv}.
\end{enumerate}
The multi-agent Gym API can be readily plugged into an RL training framework such as RLLib (grey), where agent-level policies (red) are learned. Once the individual device physics has been implemented according to the \texttt{ComponentEnv} API, the software automates the creation of \texttt{MultiComponentEnv} environments.  Any number of \texttt{ComponentEnv} and \texttt{MultiComponentEnv} agents can then be added to the \texttt{MultiAgentEnv} environment.

\subsubsection{Power Flow Solver Integration} Another key feature of PowerGridworld is the integration of a power flow solver for simulating the grid physics that underlies the multi-agent environment.  Although our examples utilize the open distribution system simulator (OpenDSS) \cite{opendss} to solve the power flow on a test feeder, any power flow solver wrapped in the \texttt{PowerFlowSolver} API can be utilized.

\subsection{Advantages}
The advantages of using PowerGridworld over MARL software packages are as follows.  First, the plug-and-play modularity with a three-tier hierarchy (\emph{cf.} Fig. \ref{fig-architecture}) allows environments to be created from simpler components. Second, the multi-agent environment design allows both homogeneous and heterogeneous agent types. Third, the power flow solution can be used in agent-level states and rewards. Finally, PowerGridworld adheres to RLLib's multi-agent API, with converters for both CityLearn/GridLearn and OpenAI's MADDPG interfaces.

\subsection{Limitations}
Next, we list some of the limitations of PowerGridworld.  First, time stepping is synchronous and of fixed frequency.  However, we have a road map for implementing both hierarchical and multi-frequency time stepping. Second, the communication model is limited. Centralized communication, whereby the process driving the environment collects and communicates variables between agents, is relatively straightforward to implement using only the Gym API.  More advanced paradigms require custom implementations. Finally, the initial version of the  \texttt{MultiAgentEnv} serializes calls to the agents (i.e., there is no parallelism).

\section{Case Studies}
In this section, we present two examples of how PowerGridworld can be used to formulate multi-agent control tasks in energy systems.

\subsection{Multi-Agent Building Coordination Environment}\label{subsec-building-coordination}

In the first example, we consider three RL agents in a homogeneous setting. Each agent controls three components within one building: one HVAC system, one on-site PV system, and one energy storage (ES) system. Using this setup, this example demonstrates how to use the PowerGridworld package to model a learning environment that allows agent coordination while achieving each agent's own objective.

To this end, the MARL system is implemented as follows:

\begin{enumerate}
\item For each agent/building, the HVAC system needs to be controlled so that thermal comfort can be realized with minimal energy consumption. As a result, the HVAC component reward, $r_t^{i, HVAC}$, includes penalties for both thermal discomfort and energy consumption.

\item The PV and ES systems are two additional components that are controlled by an agent to modify the building's net power consumption, but for simplicity, the rewards related to these two components are set to be zero, i.e., $r_t^{i, PV} = r_t^{i, ES}=0$.

\item We designed a simple scenario with a sudden PV generation drop when the system loading level is high. If all three buildings, which connect to the same bus in the distribution system, only care about their own objective, the voltage at the common bus might fall below the limit, i.e., $v_{comm} < \underline{v}$. As a result, voltage support -- maintaining $v_{comm} \geq \underline{v}$ -- requires the three buildings to coordinate with one another.
\end{enumerate}

\begin{figure}[t]
\centering
\includegraphics[width=1.0\linewidth]{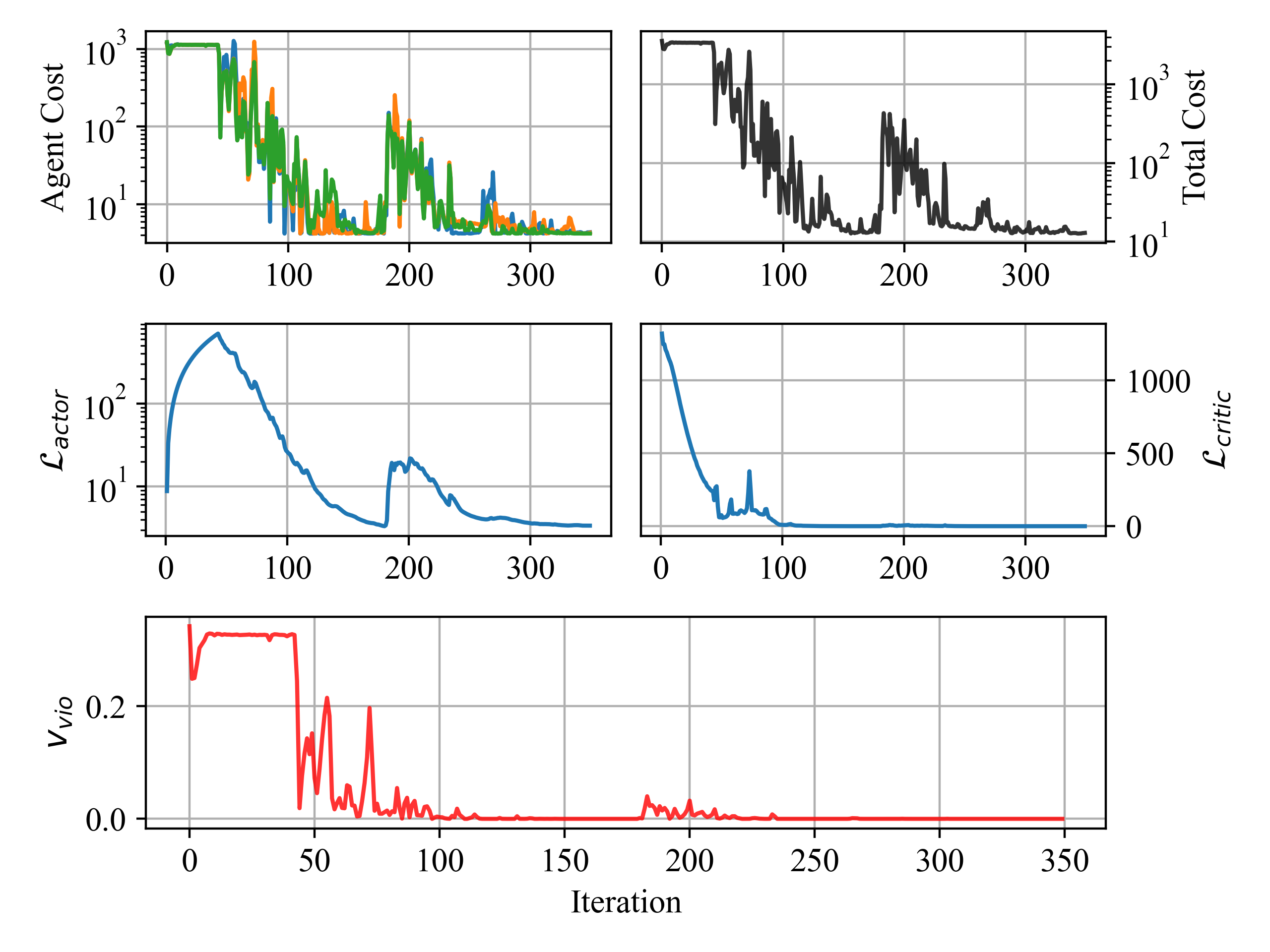}
\caption{Learning curves of using MADDPG to train control policies for the multi-agent coordinated building control. All x-axes represent training iterations. Losses, i.e., $\mathcal{L}_{actor}$ and $\mathcal{L}_{critic}$, are averaged among three agents.}
\label{fig-homo-mae}
\end{figure}

Based on the setup above, at control step $t$ and for agent $i$, the total agent reward is

\begin{equation}
    r_t^i = r_t^{i, agent} + r_t^{i, sys}
\end{equation}
in which $r_t^{i, agent}= r_t^{i, HVAC} + r_t^{i, PV} + r_t^{i, ES}$ is the agent-level reward and $r_t^{i, sys} = -\lambda [\text{max}(0, v_{comm}-\overline{v})+\text{max}(0, \underline{v}-v_{comm})] / 3$ represents the system-level reward, shared evenly with all three agents. Here, $\lambda$ is a large number. Through MARL training, each agent should be able to optimize its own objective (i.e., keep $r_t^{i, agent}$ low) and also able to work with other agents to avoid any voltage violation (i.e., keep $r_t^{i, sys}$ low).

To train control policies for this problem, we use OpenAI's MADDPG implementation. Specifically, agent $i$ trains a critic network (i.e., $Q_{\theta^i}(\mathbf{s}, \mathbf{a})$) in an off-policy manner to minimize the mean squared Bellman error (MSBE):

\begin{equation}
    \mathcal{L}_{critic}^i (\theta^i) = \mathbb{E}_{\mathbf{s}, \mathbf{a}, r^i, \mathbf{s'}}[[Q_{\theta^i}(\mathbf{s}, \mathbf{a})-(r^i + \gamma Q_{\theta^{i,-}}(\mathbf{s'}, \mathbf{a'})]^2]
\end{equation}
and the actor (i.e., the control policy $\mu_{\phi^i}(s^i)$) is trained by minimizing the following actor loss:

\begin{equation}
    \mathcal{L}_{actor}^i (\phi^i) = -\mathbb{E}_{\mathbf{s}, \mathbf{a}}[Q^i_{\theta}(\mathbf{s}, [... a^{i-1}, \mu_{\phi^i}(s^i), a^{i+1}, ...])]
\end{equation}
In the above equations, $\theta^i$ and $\phi^i$ are the RL parameters to be optimized, and $\theta^{i, -}$ represents the target value network parameters (a common off-policy learning trick; see \cite{mnih2015human} for details). In our notation, $a^i$ and $s^i$ are the action and state of agent $i$, respectively, and the collection of all agents' actions and states are written as $\mathbf{a}$ and $\mathbf{s}$. the The states at the next step are denoted $\mathbf{s'}$, and $\mathbf{a'}=[\mu_{\phi^i}(s^i) | i={1, ...}]$ are the policy chosen actions at $\mathbf{s'}$.

Fig. \ref{fig-homo-mae} shows the learning curves over 350 training iterations. By the end of the training, both agent costs and the total cost converge to a low level. $\mathcal{L}_{critic}$ starts at a large value and gradually decreases to a value close to zero, indicating that the state-action values can be estimated accurately. As the value estimation becomes more reliable, $\mathcal{L}_{actor}$ also decreases, implying that the control policies are improving to achieve a higher reward level for each agent. Finally, the episodic voltage violation sum, $v_{vio}$, is high at the beginning, and as the agents learn to coordinate with one another, the voltage violation is also eliminated, leading to $v_{vio}=0$.

In summary, this example demonstrates using PowerGridworld to formulate a MARL problem with both competition (building comfort) and collaboration (system voltage) among agents. Admittedly, instead of na\"ively splitting the system penalty evenly among agents to encourage agents' coordination, a more advanced approach could be flexibly implemented using this framework by modifying the corresponding interfacing functions.

\subsection{Multi-Agent Environment With Heterogeneous Agents}

A key feature of the PowerGridworld package is that it enables users to model heterogeneous agents that interact both with one another and with the grid.  To demonstrate this feature, we developed a simple example with three different agents consisting of one smart building---simulated as a \texttt{MultiComponentEnv} composed of a five-zone building, a solar panel, and a battery component environment---and two independent component environments representing a PV array and an electric vehicle (EV) charging station.  The agents here are loosely coupled according to their reward structures and observation spaces, as described next.

\emph{Smart building ($i=1$).} The five-zone building has a simple reward function characterized by a soft constraint that zone temperatures be maintained within a comfort range.  The building thermal model used is the same as in Section \ref{subsec-building-coordination};  the reward function is similar, except that it does not take power consumption into account. 

\emph{PV array ($i=2$)}.  Next, we include a controllable PV array as a source of real power injection, with the purpose of mitigating voltage violations stemming from high real power demand on the distribution feeder.  We model a simple control whereby the real power injection can be curtailed between 0\% and 100\% of available power from the panels; the observation space consists of both real power available from the panels and the minimum bus voltage on the feeder, $v_{min}$.  (The scenario we consider is stable with respect to maximum feeder voltage.)  The reward function is given by a soft penalty on the minimum bus voltage, which is computed using OpenDSS.

\emph{EV charging station ($i=3$)}.  Finally, we consider an EV charging station with an aggregate, continuous control, $a_3 \in [0, 1]$, representing the rate of charging for all charging vehicles.  For example, with action $a_3 = 0.25$, all charging vehicles will charge at 25\% of the maximum possible rate.  The distribution of vehicles is control dependent because, as vehicles become fully charged, they leave the station and thus reduce the aggregate load profile.  Furthermore, each vehicle has prespecified (exogenous) arrival and departure times, before and after which it cannot charge.  The observation space consists of a handful of continuous variables characterizing the station's current occupancy and aggregate power consumption, as well as aggregate information about the state of the charging vehicles. The reward function balances the local task of meeting demand with a grid-supportive task of keeping the total real power consumption under a peak threshold.
Note that, while the charging station does not directly respond to grid signals, the soft constraint on peak power incentivizes load shifting.

\begin{figure}[t]
\centering
\includegraphics[width=1.0\linewidth]{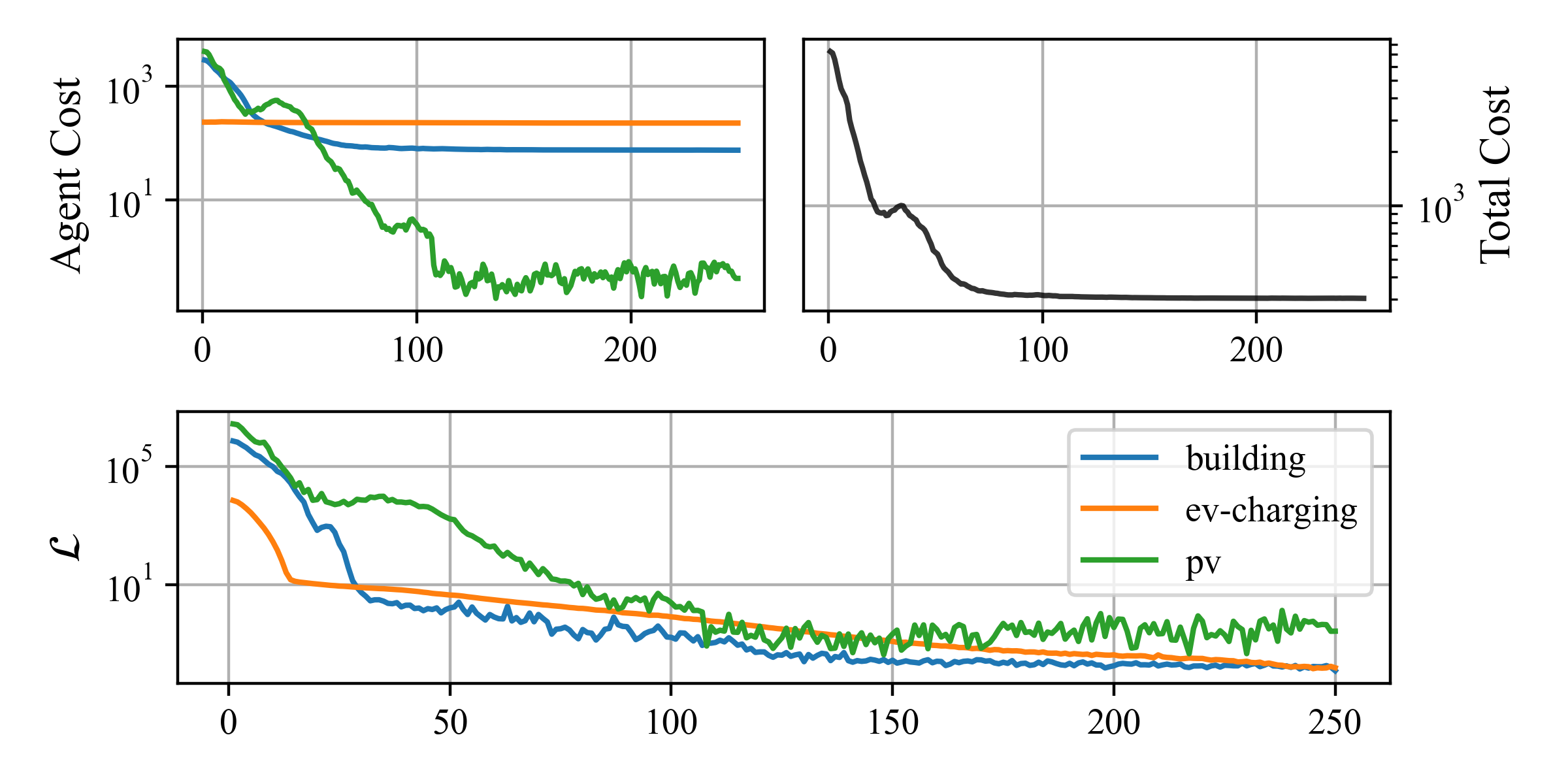}
\caption{Learning curves for independent PPO policies for the heterogeneous control problem trained using RLLib.  The PPO loss function, $\mathcal{L}$, is given in (\ref{eqn-ppo-loss}).  All x-axes represent training iterations.}
\label{fig-heterogeneous-ppo}
\end{figure}

Using RLLib's multi-agent training framework, we train separate proximal policy optimization (PPO) \cite{schulman2017proximal} policies for each agent, with each agent attempting to optimize its own reward.  Although training multi-agent policies in this way is generally challenging due to nonstationarity, here, the agents are only loosely coupled through bus voltages in the PV agent's reward function, and training converges without issue---see Fig. \ref{fig-heterogeneous-ppo}.  The lower panel in the figure shows the PPO loss function for each agent's policy,
\begin{align}
    \mathcal{L} = \mathbb{E} \Bigg[ \sum_{t=0}^T \min \left( \frac{\pi_{\theta}(a_t|s_t)}{\pi_{\theta_{old}}(a_t|s_t)}\hat{A}(a_t, s_t), g\left(\epsilon, \hat{A}(a_t, s_t)\right) \right)\Bigg]
    \label{eqn-ppo-loss}
\end{align}
where $\hat{A}$ is the advantage estimator, $g(\epsilon, \cdot)$ is a clipping function with threshold $\epsilon$, and $\theta_{old}$ refers to the policy weights from the previous training iteration.  We refer the reader to \cite{schulman2017proximal} for additional details about the PPO algorithm and loss function.

\section{Conclusion}

PowerGridworld fills a gap in MARL for power systems by providing users with a lightweight framework for rapidly prototyping customized, grid-interactive, multi-agent simulations with a bring-your-own-model philosophy.  The multi-agent Gym API and other API converters enable users to rapidly integrate with existing MARL training frameworks, including RLLib (multiple algorithms) and OpenAI's MADDPG implementation.  Unlike the CityLearn and GridLearn software packages, PowerGridworld does not provide carefully designed benchmarks for a given application, such as demand response with voltage regulation.  Rather, it provides the user with abstractions that streamline experimentation with novel multi-agent scenarios, component Gym environments, and MARL algorithms where the power flow solutions are essential to the problem.  Integration with RLLib, in particular, paves the way for the use of supercomputing and HPC resources for RL training, which will become ever more important as the complexity of MARL simulations continues to increase.

\bibliographystyle{IEEEtran}
\bibliography{reference}

\end{document}